\documentclass{article}
\usepackage{ijcai19}

\usepackage{times}
\usepackage{soul}
\usepackage{url}
\usepackage[hidelinks]{hyperref}
\usepackage[utf8]{inputenc}
\usepackage[small]{caption}
\usepackage{graphicx}
\usepackage{amsmath}
\usepackage{booktabs}
\usepackage{natbib}
\usepackage{threeparttable}
\usepackage{color}

\usepackage{multirow}
\usepackage{amssymb}
\usepackage{bm}

\usepackage{longtable}
\newcommand{\tabincell}[2]{\begin{tabular}{@{}#1@{}}#2\end{tabular}}

\renewcommand{\cite}{\citep}

\title{Exploring Unsupervised Pretraining and Sentence Structure Modelling for Winograd Schema Challenge}

\author{
Yu-Ping Ruan$^1$\and
Xiaodan Zhu$^2$\and
Zhen-Hua Ling$^{1}$\and
Zhan Shi$^2$ \and
Quan Liu$^3$ \And
Si Wei$^3$\\
\affiliations
$^1$National Engineering Laboratory for Speech and Language Information Processing,\\
University of Science and Technology of China, Hefei, P.R.China\\
$^2$Department of Electrical and Computer Engineering, Queen's University, Kingston, Canada\\
$^3$iFLYTEK Research, Hefei, P.R. China\\
\emails
yupingruan@gmail.com,
zhu2048@gmail.com, zhling@ustc.edu.cn,
18zs11@queensu.ca,
\{quanliu, siwei\}@iflytek.com
}

\begin{document}

\maketitle

\begin{abstract}
Winograd Schema Challenge (WSC) was proposed as an AI-hard problem in testing computers’ intelligence on common sense representation and reasoning. This paper presents the new state-of-the-art on WSC, achieving an accuracy of 71.1\%. We demonstrate that the leading performance benefits from jointly modelling sentence structures, utilizing knowledge learned from cutting-edge pretraining models, and performing fine-tuning. We conduct detailed analyses, showing that fine-tuning is critical for achieving the performance, but it helps more on the simpler associative problems. Modelling sentence dependency structures, however, consistently helps on the harder non-associative subset of WSC. Analysis also shows that larger fine-tuning datasets yield better performances, suggesting the potential benefit of future work on annotating more Winograd schema sentences.
%we marked down several negative results that we believe to be useful in shedding light on future study. 
\end{abstract}

\section{Introduction} 
The impressive advance achieved in the last several years on distributed representation and neural networks has resulted in significant improvement in many
research areas \cite{krizhevsky2012imagenet,mnih2013playing,mikolov2013distributed,bahdanau2014neural,vaswani2017attention,lample2017unsupervised, devlin2018bert}, which in turn triggers further curiosity to understand how much such models can further solve hard AI problems. 

Winograd Schema Challenge (WSC) was proposed as an AI-complete problem in testing computers’ intelligence on common sense reasoning \cite{levesque2012winograd,morgenstern2015winograd, marcus2016beyond}. 
WSC elegantly embeds common sense reasoning into a simple form: a binary classification on coreference resolution. A typical WSC example was proposed in \citet{winograd1972understanding}: for the sentence “\textit{The
city councilmen} refused \textit{the demonstrators} a permit because
\textit{\underline{they}} feared violence.” and a corresponding question “Who
does the word \textit{\underline{they}} refers to?”, a computer is expected to find the answer, i.e., \textit{the city councilmen} but not \textit{the demonstrators}.

While answering WSC questions is usually very simple for common human beings, it presents a great challenge for machines. Recently, there have been many different efforts attempting to solve the problem (refer to Section~\ref{sec:related} for a brief survey). 

%In the works of [Sharma, 2014; Sharma et al., 2015], they identified the knowledge needed to answer a challenge question, hunted down that knowledge from text repositories, and then made logic reasoning with them to output the answer. However, this approach relied on the WS test set to extract corresponding knowledge, which had very poor scalability. In [Schuller, 2014 ], they proposed to tackle WS problems by formalizing relevance theory in knowledge graphs. Their experiments were performed using Answer Set Programming. However, due to the complexity of the proposed method, they just examined how to answer four WS questions in that work.

%While it is hard to believe the final solution to such problems will be resulted from the current technology based on distributed representation, as reasoning is traditionally being solved with symbolic approaches and, this paper push forward the state-of-the-art along the line. 

%The unsupervised pretrained models on large text corpus have recently achieved impressive performance on many NLP tasks.
%It yet remains a basic question whether such pretrained models can learn common sense for WSC? How they help the WSC? In what way, and Why?

In this paper, %we propose a method of adding dependency into the self-attention process in Transformers to explicitly model the sentence structures and demonstrate its effectiveness on WSC. 
we present the new state-of-the-art model for WSC, achieving an accuracy of  71.1\%. We demonstrate that the leading result benefits from jointly introducing sentence structures into modelling, utilizing external knowledge learned from cutting-edge pretraining, together with fine-tuning using the \emph{Raham-Ng} dataset~\cite{rahman2012resolving}.

In addition, we conduct detailed analyses. We observed that that fine-tuning is critical for achieving the performance, but it helps more on solving the simpler associative problems \cite{trichelair2018evaluation}. Modelling sentence dependency structures, however, consistently helps on the harder non-associative subset of WSC. Analysis also shows that larger fine-tuning datasets yield better performances, suggesting the potential benefit of future work on annotating more Winograd schema sentences, which may help further complement and leverage the pretrained models.

\section{Related Work}
\label{sec:related}
Some previous efforts on resolving the Winograd Schema problem relied heavily on the annotated knowledge, hand-crafted features, and (or) rule-based reasoning \cite{peng2015solving, bailey2015winograd,schuller2014tackling, liu2017combing, liu2017cause}. In particular, \citet{rahman2012resolving} employed human annotators to build more supervised training data, in which the models utilized nearly 70 thousand hand-crafted features, including querying data from Google Search API.

More recently \citet{sharma2015towards} utilized a semantic parser on the WSC sentences, queried texts through Google Search, and reasoned on the graph produced by the parser.  \citet{emami2018knowledge} showed better performance along the same direction.
\citet{schuller2014tackling} formalized a knowledge-graph data structure and a reasoning process based on cognitive linguistics theories. \citet{bailey2015winograd} introduced a framework for reasoning using expensive annotated knowledge bases as axioms, while \citet{liu2017combing} incorporated several knowledge bases into the training process of skip-gram word embeddings, and the resulting knowledge-enhanced embeddings were used to better score the candidates. 

Unlike the above work, we are first curious about the effectiveness of cutting-edge pretrained models. We show they do help achieve the state-of-the-art performance, but we observe that they help more on the simpler associative problems. We propose to incorporate sentence structures into the pretraining and fine-tuning framework, which consistently helps on the harder non-associative subset of WSC. 

%Among different models, we found that adding dependency into transformers' self-attention yielded the best result. 

%These models have shown to be very effective on many different NLP tasks, but there is no detailed study on how they help the WSC.which, unlike the on-the-fly search based on WSC questions,  encode knowledge and common sense in the pretraining transformer-based data structures. 

%As stressed in \citet{levesque2012winograd}, the WSC problems are designed to work against techniques such as traditional linguistic restrictions, common heuristics, or simple statistics evidence from text (``Google-proof"). 

For research on unsupervised pretraining, there exists considerable work in the literature but the typical models proposed in the most recent years include GPT, ELMo, and BERT   ~\cite{radford2018improving, peters2018deep, devlin2018bert}, which achieved impressive performance on a wide variety of tasks. Among the models, we choose BERT \cite{devlin2018bert}, which is among the-state-of-art. 

%Inspired by these, \citet{trinh2018simple} propose several specially trained LMs on large corpus and use these LMs to predict the resolution of the ambiguity with higher probability. %Very recently, OpenAI published their bigger pretraind model ``GPT-2" and report a very impressive accuracy, i.e., 70.7\%, on the WSC set.

%\section{Methodology}
\section{Understanding the Roles of Pretraining and Fine-Tuning for WSC}
While the unsupervised pretrained models on large text corpus have recently achieved impressive performance on various NLP tasks,
it remains a fundamental question on how such pretrained model learn common sense to help solve hard common sense reasoning problems. For simple inference problems, BERT has achieved performance comparable to that of human being's, e.g, on the recently published \emph{SWAG} dataset \citep{zellers2018swag}. This invites a further investigation on the harder WSC problems.

%? In what way, and how much?

%The state-of-the-art of unsupervised utilizing large language data is pretrain and fune-tuning framework. It is a open question if such model learn common sense from unstructured text? How they help the WSC? In what way, and Why? 

While BERT has been tested on the GLUE dataset~\cite{wang2018glue}, there have no conclusive results on WSC due to data splitting issues~\cite{devlin2018bert}. This paper is the first to perform a detailed study on the state-of-the-art pretraining-finetuning framework for WSC. %BERT~\cite{devlin2018bert} is one of the most advanced pretrained models. 
%As one of the most advanced pretrained models, BERT has achieved impressive state-of-art results on a number of language understanding benchmarks \cite{devlin2018bert}. 

%Here we start to test the ability of pretrained BERT on the WSC, aiming to explore how pre-learned knowledge in BERT performs on the WSC.
%which is proposed as an AI-hard problem in testing computers intelligence on common sense representation and reasoning.
\paragraph{WSC as Next Sentence Prediction}
It is reasonable to utilize BERT to learn and encode common sense knowledge existing in large text corpora in an unsupervised manner. While BERT has two targets, i.e., optimizing masked language models (LMs) and next-sentence prediction (refer to \cite{devlin2018bert} for details), to solve the WSC problems, we formulate the WSC sentences in the next-sentence prediction framework; a specific example is presented as follows:

\begin{itemize}
    \item \textbf{Original WSC sentence}: \emph{The trophy} doesn't fit into \emph{the brown suitcase} because \underline{\emph{it}} is too large.
    \item \textbf{Candidate sentence 1}: \texttt{[CLS]} The trophy doesn't fit into the brown suitcase because \texttt{[SEP]} \textit{\underline{the trophy}} is too large . \texttt{[SEP]}
    \item \textbf{Candidate sentence 2}: \texttt{[CLS]} The trophy doesn't fit into the brown suitcase because \texttt{[SEP]} \textit{\underline{the brown suitcase}} is too large. \texttt{[SEP]}
\end{itemize}

In the formulation, we replace the pronoun ``\emph{it}" in the original WSC sentence with the two candidate entities ''\emph{the trophy}" and ``\emph{the brown suitcase}", respectively, to derive two candidate sentences, where a special classification token \texttt{[CLS]} and delimiter token \texttt{[SEP]} are inserted in following \citet{devlin2018bert}.

%\begin{itemize}
%    \item \textbf{Original sentence}: \emph{The trophy} doesn't fit into \emph{the brown suitcase} because [it] is too large.
%    \item \textbf{Candidate sentence1}: \emph{The trophy} doesn't fit into \emph{the brown suitcase} because \textbf{\emph{the trophy}} is too large.
%    \item \textbf{Candidate sentence2}: \emph{The trophy} doesn't fit into \emph{the brown suitcase} because \textbf{\emph{the brown suitcase}} is too large.
%\end{itemize}

%In the formulation, we replace the pronoun ``\emph{it}" in the original sentence with the two candidate entities ''\emph{the trophy}" and ``\emph{the brown suitcase}", respectively to derive two candidate sentences. \xd{how to split exactly?} The final sentence pair as inputs to BERT is as follows :

%\begin{itemize}
%   \item \textbf{Input1}: [CLS] The trophy doesn't fit into the brown suitcase because [SEP] the trophy is too large . [SEP]
%    \item \textbf{Input2}: [CLS] The trophy doesn't fit into the brown suitcase because [SEP] the brown suitcase is too large . [SEP]
%\end{itemize}
Then we score these two candidate sentences using pretrained next-sentence-prediction. The substitution that results in a more probable candidate sentence will be the correct answer. 

%This manner is similar with that in \citet{trinh2018simple}, in which the pretrained language models are used to score how probable the part of the word sequences following the candidates.

\paragraph{Fine-tuning BERT for WSC}
%Based on that fine-tuned BERT has achieved impressive results on several language understanding benchmarks, it would be interesting and instructive to study whether fine-tuning would benefits BERT's performance on the AI-hard problem, i.e., WSC set.
As we will discuss later, we find fine-tuning is critical for WSC. The original WSC dataset has only 273 sentences. We did not observed any performance gain using the 273 sentences to fine-tune BERT with 10-fold cross validation.
It is therefore interesting to understand if more Winograd schema sentences will benefit the framework. Fortunately, \citet{rahman2012resolving} gathered a dataset consisting of $1,886$ sentences of pronoun resolution problems\footnote{available at \url{http://www.hlt.utdallas.edu/~vince/data/emnlp12/}}. To ensure that there is no overlapping sentences between this dataset and the WSC dataset, we manually investigated and removed from the $1,886$ sentences $4$ sentences that are overlapping with the WSC set. The remaining $1,882$ sentences, referred to as the \emph{Raham-Ng} dataset in the remainder of this paper, will be used to fine-tune the pretrained BERT models to develop our state-of-the-art models and  analyze how pretraining and fine-tuning help WSC.
%Luckily, \citet{rahman2012resolving} gathered a dataset containing $1,886$ sentences of similar pronoun resolution problems (referred to later as  the \emph{Raham-Ng} dataset\footnote{available at \url{http://www.hlt.utdallas.edu/~vince/data/emnlp12/}}), following \citet{winograd1972understanding} and \citet{levesque2012winograd}, which can be used to fine-tune BERT. To make sure that there is no overlapping or similar sentences between the \emph{Raham-Ng} and the WSC set, we calculate the words overlapping rates between sentences in the \emph{Raham-Ng} and those in WSC, and remove sentences in \emph{Raham-Ng} whose words overlapping rate with any one certain sentence in the WSC bigger than  58\%\footnote{We select this overlapping rate threshold by human checking whether the filtered sentence from the \emph{Raham-Ng} and the corresponding sentence from the WSC are truly coincident concerning their internal common sense rationale, and we decide 58\% can assure this.}. After that, only 4 sentence in \emph{Raham-Ng} are filtered, and the remaining $1,882$ sentences, denoted as \emph{Raham-Ng}, are used for our subsequent experiments.
\begin{table*}[!t]
\centering
\begin{tabular}{l|p{10.0cm}l}
\hline
Sentence Type  & Examples & Proportion \\
\hline
Associative & \tabincell{l}{\emph{Fred and Alice} had very warm down \emph{coats}, but \textit{\underline{they}} were not prepared \\for the cold in Alaska. \\[9pt] I'm sure that my \emph{map} will show this \emph{building}; \textit{\underline{it}} is very good.}  & 13.5\% (37)  \\[16pt]
\hline
Non-Associative & \tabincell{l}{\emph{The large ball} crashed right through \emph{the table} because \textit{\underline{it}} was made \\ of steel. \\[9pt] \emph{The firemen} arrived after \emph{the police} because \textit{\underline{they}} were coming from\\ so far away.}  & 86.5\% (236) \\[16pt]
\hline
\end{tabular}
\caption{Examples and proportions of associative and non-associative instances in the WSC.}
\label{tab:example_ass}
\end{table*}
\section{Modelling Sentence Structures} \label{sec:add_dep}
In addition to investigating common sense knowledge learned with pretraining and encoded in the Transformer-based data structures, we believe carefully modelling WSC sentences themselves are important, as human rely heavily on sentence syntax and semantics to solve the WSC problems. Unlike previous models~\cite{liu2017combing,sharma2015towards}, in this paper we explore modelling sentence structures together with the pretraining-finetuning framework, to leverage and combine their complementary strengths. 

%have started to consider some stru to find external knowledge from the Web and semanti. 

BERT is mainly built on deep-stacked bidirectional Transformers \cite{vaswani2017attention}. A typical Transformer works purely on attention mechanism and does not have an explicit notion of word order beyond marking each word with its absolute position embedding. It has been found that the RNNs are superior to the full attention network, i.e, Transformer, on modelling some hierarchical structures in sentences \cite{tran2018importance}. In this paper we will explicitly incorporate dependency structure into BERT. For many tasks that are sensitive to sentence structures where a relatively small training data are provided, an explicit utilization of sentence structures often benefits. We wonder if WSC can benefit from both sentence structure and pre-learned knowledge jointly, and if so how?

%Following parts will have a brief introduction to the self-attention in Transformer and how to add dependency into BERT.
\begin{figure}[!t]
	\centering
	\includegraphics[width=2.5in]{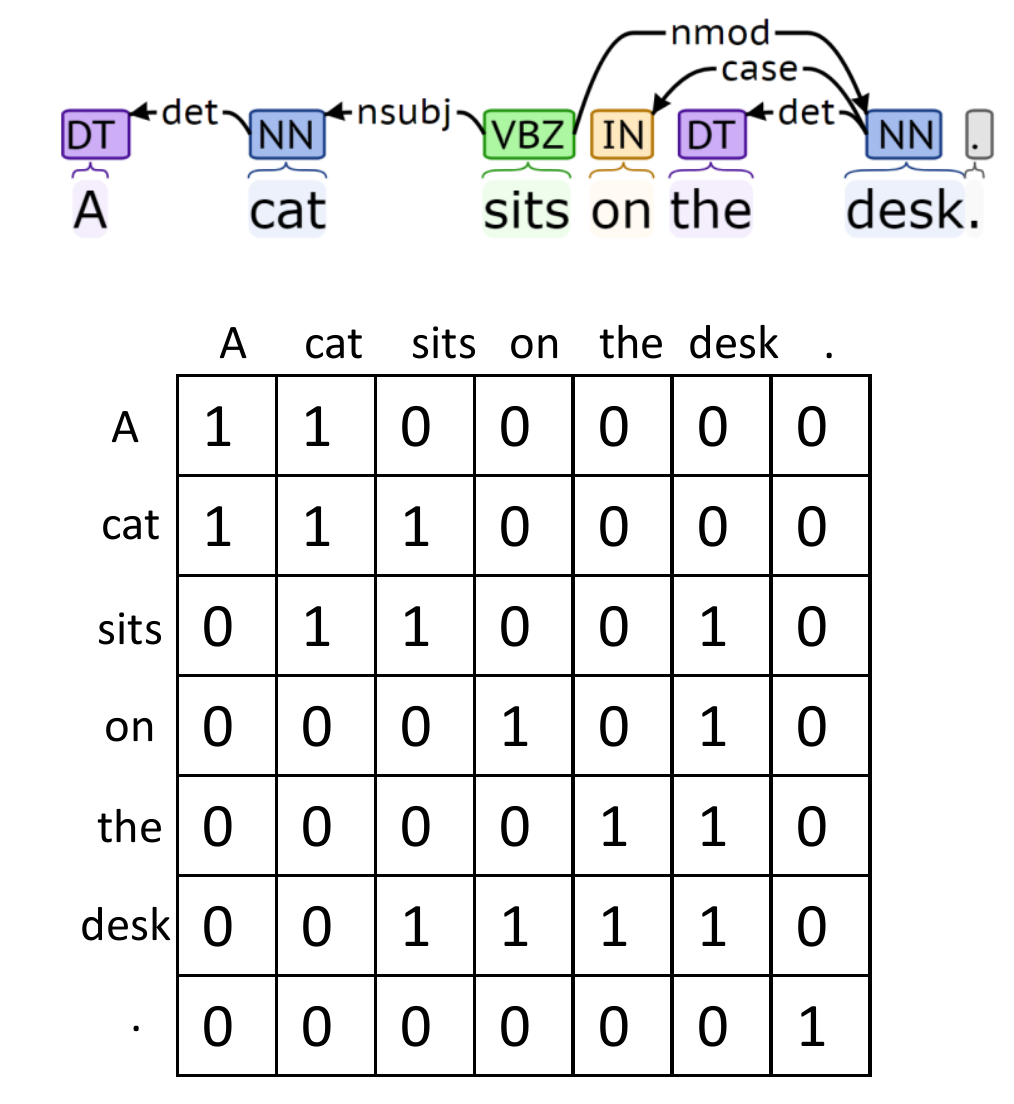}
	\caption{Dependency parsing of a sentence ``A cat sits on the desk.", and the corresponding dependency mask matrix.}
	\label{fig:dep_tree}
\end{figure}
\paragraph{Adding Dependency into Transformers}  Transformers in BERT consist of multiple layers \cite{vaswani2017attention}, among which the multi-head self-attention layer serves as the most important component. The self-attention layer in Transformer is presented as following, while the detailed discussion can be found in \cite{vaswani2017attention}.

For input hidden states $H = [\mathbf{h}_1, \mathbf{h}_2, ..., \mathbf{h}_n]$ corresponding to $n$ tokens in the sequence, which can be the output of last Transformer layer or the embedding layer, one self-attention head, i.e., ${head}_i$, derive its output $H^{'(i)}$ as follows:
\begin{align}
   \label{eq:eq1} Q^i &= HW^Q_i,\\[1mm]
   \label{eq:eq2} K^i &= HW^K_i,\\[1mm]
   \label{eq:eq3} V^i &= HW^V_i,
\end{align}
where $Q^i$, $K^i$, $V^i$ represents the query, key, value matrices respectively, and $W^Q_i$, $W^K_i$, $W^V_i$ are three projection matrices. 

Then a dot-product attention and softmax-weighted sum is applied to $Q^i$, $K^i$, $V^i$ to derive the output $H^{'(i)}$:
\begin{align}
   \label{eq:eq4} A^i &= \frac{Q^i{K^i}^T}{d_k}, \\[1mm]
    H^{'(i)} &= softmax(A^i) V^i, 
\end{align}
where $d_k$ is the scaling factor. Finally, outputs $\{H^{'(i)}\}$ from multiple heads will be concatenated to form the output of the self-attention layer.

%, which may weakening the utilization of hierarchical structure existing in sentences such as dependency among words

For the original Transformers in BERT, as illustrated above, each word will attend to all words in the sequence when in the process of self-attention. To add explicit structural constraints, we propose to combine the dependency structures with self-attention in Transformer. 

%To explicitly adding dependency structure into BERT but not change BERT too much, we adopt to use a dependency mask to control each word only attend to its head word, child words, and itself. 

We propose a Dependency Mask technology to constrain a word to specially attend to its head word, child words, and itself. As shown in Figure \ref{fig:dep_tree}, for sentence ``A cat sits on the desk.", we first derive its dependency tree and then convert it to a simpler dependency mask matrix, denoted as $D$ here, each row $\mathbf{d}_i$ in $D$ represents the relation between the $i$-th word and all words in the sentence, the positions indicating its head word, child words, and itself are set to $1$. The dependency mask $D$ are added to the softmax-weighted sum as follows:
\begin{align}
    H^{'(i)} &= softmax(A^i \times D) V^i, 
\end{align}
where $A^i$ and $V^i$ are the same with that in Eq. \ref{eq:eq4} and Eq. \ref{eq:eq3}, the $D$ is used to mask the dot-attention weights $A^i$.

%\paragraph{Where and How to Add the Dependency into BERT} 

%As a complete coordinated whole as itself,  BERT is powerful but also subtle, directly adding dependency into internal layers in BERT may harm its performance. 

%BERT is pretrained on large text corpora and composed of multiple Transformer layers. 

We explore two approaches to adding the dependency structures into BERT: the \textit{inside} and \textit{outside} approach.

\begin{itemize}
    \item \emph{Inside}: In the inside approach, the dependency mask is added to Transformer layers inside BERT---we mask the attention weights in the last, middle, or first $t$ Transformer layers.
    \item \emph{Outside}: While BERT is powerful, it may be very sensitive to the modification that is made only in the fine-tuning phase (but not in pretraining), particularly directly in the internal layers. Motivated by that, we also propose the \textit{outside} incorporation. Specifically we define $t$ layers of new Transformers, denoted as TransformerRNN-$t$ here, upon the last Transformer layer of the original BERT model. These new layers share the same set of parameters. The TransformerRNN-$t$ works in a recurrent manner, motivated by the universal Transformers \cite{dehghani2018universal}. With this, the dependency mask is added to all layers in TransformerRNN-$t$.
\end{itemize}

\section{Evaluation Protocols}\label{sec:eva_protocol}
Due to the difficulty to acquire high-quality samples for common-sense reasoning under the Winograd Schema settings, the WSC dataset comprises only 273 test instances.\footnote{Recently, 12 new sentences have been added.} Previous state-of-the-art on the full WSC set for \emph{single-model} performance is around 55\% \citep{trinh2018simple,emami2018knowledge}. Indeed, it has been found that there is more than a 1-in-3 chance of scoring above 55\% accuracy with a set of random classifiers \citep{trichelair2018evaluation}. So achieving above random accuracy on the WSC does not necessarily corresponding to success on common sense reasoning. In particular, \citet{trichelair2018evaluation} proposed two new evaluation protocols to alleviate above evaluation problems by subdivide and augment the WSC data according to two properties: \emph{associativity} and \emph{swithchability}. 

In this paper we use these effective protocols to help probe the roles of pretraining-funetuning frameworks and our proposed models in WSC. 

%For example, we show that fine tuning is critical but is helpful more on the simple associative subset while the incorporation of sentence dependency consistently help more on the non-associative sentences.

\paragraph{Associativity}
As one of the designing principle \cite{levesque2012winograd}, the sentence in WSC should not be resolvable via simple statistics that associate a candidate antecedent to  certain components in the sentences. For example, in the above formulation, the statement ``The \emph{lions} ate the \emph{zebras} because \textit{\underline{they}} are predators" \cite{rahman2012resolving}, we have two pairs of candidate sentences ``The \emph{lions} ate the \emph{zebras} because \textit{\underline{lions}} are predators" and ``The \emph{lions} ate the \emph{zebras} because \textit{\underline{zebras}} are predators". Such sentences can be resolved based on a much stronger association/collocation of lions with predators than that of zebras. ~\citet{trichelair2018evaluation} released a dataset in which the sentences in WSC are manually annotated either as \textit{associative} or \textit{non-associative}. Table \ref{tab:example_ass} lists some examples and the size of these two subsets.

\begin{table*}[!t]
\centering
\begin{tabular}{l|c|cc|ccc}
\toprule[1pt]
  \multicolumn{1}{c|}{\multirow{2}{*}{\emph{}}}
   & \bf{Full WSC} & \bf{Associative} & \bf{Non-Associative} & \bf{Unswitched} & \bf{Switched} & \bf{Consistent} \\
   & (\#273) & (\#37) & (\#236) & (\#131) & (\#131) & (\#131)\\
  \toprule[1pt]
   & \multicolumn{6}{c}{\multirow{1}{*}{\bf{Previous state-of-art models}}}\\
   \hline
      Single LM & 54.8\% & 73.0\% & 51.7\% & 55.0\% & 54.2\% & 31.0\% \\
      Ensemble 14 LMs & 63.7\% & 83.8\% & 60.6\% & 63.4\% & 53.4\% & 28.0\% \\ 
      Knowledge Hunter & 57.1\% & 50.0\% & 58.3\% & 58.8\% & 58.8\% & 52.9\% \\
    %   GPT-2 (117M params)$^*$ & $\approx$62.0\% & -- & -- & -- & -- & --\\
    %   GPT-2 (345M params)$^*$ & $\approx$63.0\% & -- & -- & -- & -- & --\\
      %GPT-2 & 70.7\% & -- & -- & -- & -- & --\\
  \toprule[1pt]
   & \multicolumn{6}{c}{\multirow{1}{*}{\bf{BERT without fine-tuning}}}\\
   \hline
   BERT-base & 52.0\% & 56.8\% & 51.3\% & 51.9\% & 55.0\% & 13.7\% \\
   BERT-large & 52.0\% & 48.6\% & 52.5\% & 52.7\% & 54.2\% & 22.7\% \\
   BERT-base + dependency & 52.7\% & 59.5\% & 51.7\% & 51.9\% & 54.2\% & 12.2\% \\
   BERT-large + dependency & 52.7\% & 48.6\% & 53.4\% & 54.2\% & 56.5\% & 25.2\% \\
   \toprule[1pt]
   & \multicolumn{6}{c}{\multirow{1}{*}{\bf{BERT with fine-tuning}}}\\
   \hline
   BERT-base & 64.5\% & 81.1\% & 61.9\% & 63.4\% & 64.9\% & 53.4\% \\
   BERT-large & 68.1\% & 75.7\% & 67.0\% & 70.2\% & 71.8\% & 64.9\% \\
   BERT-base + dependency & 67.4\% & 78.4\% & 65.7\% & 66.4\% & 61.8\% & 53.4\% \\
   BERT-large + dependency & \bf{71.1\%} & \bf{81.1\%} & \bf{69.5\%} & \bf{74.1\%} & \bf{72.5\%} & \bf{66.4\%}  \\
%   \hline
  \toprule[1pt]
\end{tabular}
\caption{Performance of different models on the full WSC dataset and subsets evaluated with different protocols.}\label{tab:overall_results}
\end{table*}

\paragraph{Swithchability}
A switchable sentence in WSC~\cite{trichelair2018evaluation}  means that switching the two antecedents does not obscure the sentence nor affect the rationale to make the resolution decision. A typical example from the WSC is as follows:
\begin{itemize}
    \item \textbf{Original sentence}: \emph{Paul} tried to call \emph{George} on the phone, but [Paul/George] wasn't successful.
    \item \textbf{Switched sentence}:  \emph{George} tried to call \emph{Paul} on the phone, but [Paul/George] wasn't successful.
\end{itemize}
When switching the antecedents \emph{Paul} and \emph{George}, the correct answer changes from \emph{Paul} to \emph{George} as well. A system that can correctly resolves both the original and the switched sentence indicates it learns the reasoning better than a system that is confused by the switching. The switchable subset contains 131 instances, which accounts for 47\% of the original WSC set \cite{trichelair2018evaluation}. For the evaluation, we report not only accuracy on the original unswitched and switched sentences in switchable subset but also consistent accuracy. Our consistent accuracy is computed as the number of correctly answered pairs (i.e., correctly answered WSC sentences both before and after a switch) divided by the total number of switchable sentences (i.e., 131), to reflect the absolute differences of systems' performance on the switchable subset. 

%answered sentences before the switch. Instead, we use the total 131 instances as the denominator, so the value is smaller but it also consider the models ability in that will not affect the rank of systems. 

%The reason we use this is which is referred as consistent accuracy in this paper. 

%consistency score in \citet{trichelair2018evaluation}, but use the consistent accuracy here.

%~\footnote{We do not adopt the consistency score in \citet{trichelair2018evaluation}, but use the consistent accuracy here.} on both original and corresponding switched sentences.

\section{Experiment Results and Analysis}

\subsection{Baselines \& Training Details}
We use two state-of-the-art systems as our baselines:
\begin{itemize}
    \item \textbf{Pretrained LMs}: These are specially trained and ensembled language models (LMs) \cite{trinh2018simple}, in which the language models are used to score the two sentences obtained by replacing the pronoun with the two candidate entities. The sentence that is assigned a higher probability is chosen as the answer. We will use in our experiments the best \emph{Single LM} and \emph{ensembling of 14 LMs} from \citet{trinh2018simple}.
    \item \textbf{Knowledge Hunter}: The Knowledge Hunter is a rule-based system that uses search engines to gather information for the candidate resolutions and then reasons over the gathered knowledge without relying on the entities themselves \cite{emami2018knowledge}.
\end{itemize}

\paragraph{Training Details} All $1,882$ sentences in the \emph{Raham-Ng} dataset are used as training set to fine-tune the pretrained BERT. The dependency parsing for the sentences are conducted by using spaCy tool\footnote{https://spacy.io/usage/linguistic-features\#section-dependency-parse}. Concerning the final dependency mask matrix input to the BERT, for the subword after tokenization, we just duplicate its original word's dependency relation vector $\mathbf{d}_i$ as its own. And for the special token \texttt{[CLS]} and \texttt{[SEP]} we set all elements in their corresponding $\mathbf{d}_i$ to $1$.
 
For the hyperparameters during the fine-tuning process, we use most default settings. Specifically, the learning rate is $2e-05$. The batch size for BERT-base and BERT-large are set to $16$ and $2$ respectively. The warmup rate for BERT-base is $0.5$ and that for BERT-large is $0.7$. The max sequence length is set to $128$, max training epochs is $15$. Dropout rates for both BERT-base and BERT-large are set $0.1$. We use the PyTorch implementation of BERT with the pretrained model files \texttt{bert-base-uncased} and \texttt{bert-large-uncased} provided by Google.\footnote{\url{https://github.com/huggingface/pytorch-pretrained-BERT\#Fine-tuning-with-BERT-running-the-examples}}

\subsection{Overall Performance}
Table \ref{tab:overall_results} shows the overall performances of different models. The results of the baselines (Single LM, Ensemble 14 LMs, and Knowledge Hunter) are copied from \citet{trichelair2018evaluation} and the consistent accuracies are computed from the consistency scores in~\citet{trichelair2018evaluation}. Together with these baselines, we present the performances of different models with regard to whether adopting fine-tuning or not, across different evaluation protocols and subsets. Note that BERT-base and BERT-large are pretrained on the same corpus but with different model sizes: BERT-base has 110M parameters and BERT-large has 348M. \footnote{Note that when we submit the paper, the GPT-2 (\textit{https://blog.openai.com/better-language-models/}) has just been posted, which is pretrained on much larger text corpora and has much more parameters than BERT-large; however, the corresponding code and model have not been fully released. }

We can see that our proposed model that leverages dependency structures with the fine-tuned BERT-large framework achieves the best performance on all metrics and across different evaluation protocols. Particularly it achieves the new state-of-the-art accuracy, 71.1\%, on the full WSC dataset. The detailed analysis of adding dependency information will be discussed later in this section.

We also observe that the pretraining frameworks (BERT-large) with proper fine-tuning (using the \emph{Raham-Ng} dataset) achieves an accuracy of  68.1\%, which outperforms all previous (baseline) models already. Fine-tuning plays a critical role in achieving the performance, and it helps significantly more on the associative subset. We also see that incorporating dependency structures consistently helps on the harder non-associative subset. 
%Note that BERT-base and BERT-large are pretrained on the same corpus: BookCorpus (800M words) and Wikipedia (2500M words), with different model sizes: BERT-base has 12 layers Transformers and total 110M parameters, and BERT-large has 24 layers of Transformers, larger hidden layer sizes, more self-attention heads, and in total 348M parameters.
%Followings are more detailed analyses:
\paragraph{Effects of Fine-Tuning} We can find that all BERT models without fine-tuning perform worse than all previous state-of-the-art models, but have a significant performance improvement after fine-tuning (i.e., about $10\%$-$20\%$ accuracy improvement on the full WSC set), which shows that pure pretrained BERT models are not competent on WSC, while fine-tuning is critical, which combines knowledge from Winograd annotation and that from pretrained models.
%fine-tuning on Winograd sentences can effectively teach BERT to how to utilize its pre-learned knowledge for the WSC problem.
\paragraph{Effects of Adding Dependency} Both fine-tuned BERT-base and BERT-large models benefit from leveraging dependency structures and achieve an accuracy improvement for about $3\%$ on the full WSC set, which shows the effectiveness of incorporating sentence structures into BERT for the WSC problem.
\paragraph{Effects of Pretrained Model Sizes} As BERT-large has a bigger model size than BERT-base, it can potentially accommodate to encode more knowledge during the pretraining process, with its advantages having been shown in many recent research efforts. Specifically for WSC, Table \ref{tab:overall_results} shows that fine-tuned BERT-large models significantly outperform the corresponding fine-tuned BERT-base models on across all metrics and evaluation protocols.
\paragraph{Associative v.s. Non-Associative} As discussed before, the associative WSC sentences are likely to be resolved by utilizing statistics such as co-occurency found in large text corpora. However, the non-associative WSC sentences are much more challenging. 

From the results of BERT models in Table \ref{tab:overall_results} and as we have already highlighted above, the associative sentences benefit more from the fine-tuning strategy than non-associative sentences. 
We due this to the capability of pretraining and fine-tuning frameworks in capturing simple statistics to judge that "lions are predators" is more plausible than "zebras are predators", as in the example we discussed earlier. However, such statistics seems to be less effective for solving harder non-associative problems.

%We due this to the fact that the BERT models are pretrained on large text corpora and the fine-tuning process can guide BERT to utilize the learned statistics to resolve the associative sentences. 

With the associative knowledge being effectively captured in the pretraining-finetuning mechanism, dependency knowledge can further help solve the harder non-associative problems consistently. 

%that it even sometimes harm BERT's performance on the associative sentences, but all BERT models has a significant performance improvement on non-associative sentences when utilizing dependency structure, which indicates that the dependency structure information in sentences can help to better resolve the non-associative instances in WSC.

\paragraph{Switchable \& Consistency} For the results on switchable subset in WSC,  from Table \ref{tab:overall_results}, we can observe that all models have significant performance drops on the consistent accuracy, compared to performance on the unswitched accuracy. Among these, the fine-tuned BERT-large models and Knowledge Hunter behave the most stable on the switchable subset and decrease by less than $8\%$ from the unswitched accuracy to consistent accuracy, which demonstrates that the BERT-large models do not only outperform all other models, but also have robustness comparable to the rule-based Knowledge Hunter.
\begin{table*}[!t]
\centering
\begin{tabular}{c|c|cc|ccc}
\toprule[1pt]
   
  \multicolumn{1}{c|}{\multirow{2}{*}{\tabincell{l}{Percentage of\\trainin data}}}
   & \bf{Full WSC} & \bf{Associative} & \bf{Non-Associative} & \bf{Unswitched} & \bf{Switched} & \bf{Consistent} \\
   & (\#273) & (\#37) & (\#236) & (\#131) & (\#131) & (\#131)\\
  \toprule[1pt]
   0\% & 52.7\% & 48.6\% & 53.4\% & 54.2\% & 56.5\% & 25.2\% \\
   20\% & 61.9\% & 67.6\% & 61.0\% & 64.1\% & 59.5\% & 46.6\% \\
   40\% & 65.2\% & 73.0\% & 64.0\% & 66.4\% & 64.1\% & 57.3\% \\
   60\% & 66.7\% & 70.3\% & 66.1\% & 68.7\% & 70.2\% & 63.4\% \\
   80\% & 68.5\% & 67.6\% & 68.6\% & 71.0\% & 67.9\% & 61.1\% \\
   100\% & 71.1\% & 81.1\% & 69.5\% & 74.1\% & 72.5\% & 66.4\%  \\
   \hline
  \toprule[1pt]
\end{tabular}
\caption{Performance of fine-tuned BERT-large + dependency when gradually increasing randomly sampled fine-tuning data.}\label{tab:train_size}
\end{table*}

\begin{table}[!t]
    \centering
    \begin{tabular}{l|c}
    \toprule[1pt]
      & Accuracy \\
      \toprule[1pt]
      BERT-base & \textit{64.5\%} \\
      \quad + mask first 5 layers (0-4)  & 63.7\% \\
      \quad + mask middle 5 layers (3-7)  & 67.0\% \\
      \quad + mask last 5 layers (7-11)  & \bf{67.4\%} \\
      \quad + mask all 12 layers (0-11)  & 63.7\% \\
      \quad $-----------$ & $---$ \\
      \quad + TransformerRNN-$2$ & 61.5\% \\
      \quad + TransformerRNN-$3$ & \bf{65.6\%}\\
      \quad + TransformerRNN-$5$ & 61.5\%\\
      \quad + TransformerRNN-$8$ & 58.2\%\\
      \toprule[1pt]
      BERT-large & \textit{68.1\%} \\
      \quad + mask first 5 layers (0-4)  & 67.8\% \\
      \quad + mask middle 5 layers (10-14)  & 65.6\% \\
      \quad + mask last 5 layers (19-23)  & \bf{71.1\%} \\
      \quad + mask all 24 layers (0-23)  & 66.3\% \\
      \toprule[1pt]
    \end{tabular}
    \caption{Performance of different settings for incorporating dependency into BERT with fine-tuning.}
    \label{tab:mask_settings}
\end{table}
\subsection{Detailed Analysis}

\paragraph{Adding Dependency into BERT} 
As discussed in Section \ref{sec:add_dep}, we explore two approaches to leveraging the dependency structure into pretrained models, i.e., the \emph{inside} and \emph{outside} modelling. Specifically, for the \emph{inside} method, we show here the performances of different ways to add the dependency information: masking the first, middle, last 5 layers, or masking all Transformer layers. For the \emph{outside} approach, we also show results of different settings: TransformerRNN-$2$, TransformerRNN-$3$, TransformerRNN-$5$, and TransformerRNN-$8$ respectively.

We fine-tune the BERT models on the \emph{Raham-Ng} dataset and these results are present in Table \ref{tab:mask_settings}. For the \emph{inside} method, we can see that both BERT-base and BERT-large with masking the last 5 layers achieve the best accuracy, and all other masking settings except masking middle 5 layers do not improve the performance for both BERT-base and BERT-large. For the \emph{outside} approach, we can find that only TransformerRNN-$3$ benefits the BERT-base's performance on the WSC, which is still inferior to the best setting of \emph{inside} manner. We do not experiment the \emph{outside} method on the BERT-large model due to its poor performance on BERT-base and the much larger model size of BERT-large.
%So for all other comparison experiments, we use masking last 5 Transformer layers as default settings to adding dependency structure to BERT models. 

% The dependency structure is added into BERT in a form of masking the attention weights in the self-attention layer in Transformer, considering that there are multiple Transformer layers in both BERT-base and BERT-large, we test several different settings to find the potential best way to adding the dependency into BERT: a) mask the first 5 Transformer layers, b) mask the middle 5 Transformer layers, c) mask the last 5 Transformer layers, d) mask all Transformer layers. We fine tune the BERT models on \emph{Raham-Ng} and these results are present in Table \ref{tab:mask_settings}. We can find that both BERT-base and BERT-large with masking the last 5 layers achieved best accuracy, and all other masking settings except masking middle 5 layers harm the performance for both BERT-base and BERT-large. So for the following experiments, we use masking last 5 Transformer layers as default settings to adding dependency structure to BERT models.

\paragraph{The Effects of Fine-Tuning Data Sizes}
%As emphasized in \citet{levesque2012winograd}, the WSC problem should not be resolvable via statistical evidence on large text corpora. However, o

%, which seemed to suggest that the pretrained BERT can truly have the potential ability to perform common sense reasoning on WSC

Our experiments have shown that pretrained BERT models can obtain the state-of-the-art performances on WSC with fine-tuning on a relative small dataset, i.e., the \emph{Raham-Ng} dataset. A natural question is that how the sizes of fine-tuning data affect the performances. 

We randomly selected different percentages of data from the \emph{Raham-Ng} dataset to fine-tune our best ``BERT-large + dependency" model and test their performances on the WSC dataset. The results are presented in Table \ref{tab:train_size}, which shows that a larger tuning dataset yields a better performance, suggesting the potential benefit of future work on annotating more Winograd schema sentences.

%However, results in Table \ref{tab:train_size} also shows that the performance gain gradually become smaller, which means that there may be an upper performance bound with adding more fine-tuning data.

\section{Conclusions and Discussion}
We report the new state-of-the-art performance, a 71.1\% accuracy, on the Winograd Schema Challenge (WSC). The proposed state-of-the-art solver for WSC benefits from jointly modelling sentence structures, utilizing knowledge learned from cutting-edge pretraining models, and performing proper fine-tuning. We conduct detailed analyses, showing that fine-tuning is critical for achieving the performance, but it helps more on the simpler associative problems. Modelling sentence dependency structures, however, consistently helps on the harder non-associative subset of WSC. Analysis also shows that larger fine-tuning datasets yield better performances, suggesting the potential benefit of future work on annotating more Winograd schema sentences. Although this work focuses on exploring distributed representation for WSC, caution should  certainly be taken on if that by itself will result in a final solution. 
%Among different models, we found that adding dependency into transformers' self-attention yielded the best result.   

\bibliographystyle{named}
\bibliography{ijcai19}
\end{document}